\documentclass[11pt]{article}
\usepackage{graphicx}
\usepackage{xcolor}
\usepackage{paralist}
\usepackage[compact]{titlesec}
\titlespacing*{\section}
{0pt}{5.5ex plus 1ex minus .2ex}{4.3ex plus .2ex}
\titlespacing*{\subsection}
{0pt}{5.5ex plus 1ex minus .2ex}{4.3ex plus .2ex}
\usepackage[]{ACL2023}

\usepackage{times}
\usepackage{latexsym}
\usepackage{amsmath,amssymb}

\usepackage[T1]{fontenc}

\usepackage[utf8]{inputenc}

\usepackage{microtype}

\usepackage{inconsolata}
\usepackage[labelfont=bf]{caption}
\usepackage{todonotes}

\title{Adversarial Training for Low-Resource Disfluency Correction}

\author{Vineet Bhat, Preethi Jyothi, Pushpak Bhattacharyya\\[1ex]
Indian Institute of Technology Bombay, India \\
\texttt{vineetbhat2104@gmail.com, pjyothi@cse.iitb.ac.in, pb@cse.iitb.ac.in}
}

\begin{document}
\maketitle
\begin{abstract}


Disfluencies commonly occur in conversational speech. Speech with disfluencies can result in noisy Automatic Speech Recognition (ASR) transcripts, which affects downstream tasks like machine translation. In this paper, we propose an adversarially-trained sequence-tagging model for Disfluency Correction (DC) that utilizes a small amount of labeled real disfluent data in conjunction with a large amount of unlabeled data. We show the benefit of our proposed technique, which crucially depends on synthetically generated disfluent data, by evaluating it for DC in three Indian languages- \textit{Bengali, Hindi}, and \textit {Marathi} (all from the Indo-Aryan family). Our technique also performs well in removing \textit{stuttering disfluencies} in ASR transcripts introduced by speech impairments. We achieve an average 6.15 points improvement in F1-score over competitive baselines across all three languages mentioned. To the best of our knowledge, we are the first to utilize adversarial training for DC and use it to correct stuttering disfluencies in English, establishing a new benchmark for this task.

\end{abstract}

\section{Introduction}

Disfluencies are words that are part of spoken utterances but do not add meaning to the sentence. Disfluency Correction (DC) is an essential pre-processing step to clean disfluent sentences before passing the text through downstream tasks like machine translation (\citealp{rao-etal-2007-improving}; \citealp{5494999}). Disfluencies can be introduced in utterances due to two main reasons: the conversational nature of speech and/or speech impairments such as stuttering. In real-life conversations, humans frequently deviate from their speech plan, which can introduce disfluencies in a sentence (\citealp{dell_disfluency}). Stuttering speech
consists of involuntary repetitions or prolongations of  syllables which disturbs the fluency of speech. 

Conversational disfluencies occur once every 17 words (\citealp{bortfeld_disfluency_2001}) whereas a 2017 US study\footnote{\protect\url{https://www.nidcd.nih.gov/health/stuttering}} shows that roughly 1\% of the population stutters and predominantly consists of children. One out of every four children continues to suffer from this disorder lifelong.  When such speech passes through an ASR system, readability of the generated transcript  deteriorates due to the presence of disfluencies in speech (\citealp{jones_asr}).

\citeauthor{Shriberg1994PreliminariesTA} (1994)
defines the surface structure of disfluent utterances as a combination of reparandum, interregnum and repair. The reparandum consists of the words incorrectly uttered by the speaker that needs correction or complete removal. The interregnum acknowledges that the previous utterance may not be correct, while repair contains the words spoken to correct earlier errors. 

\begin{table}[h]
        \centering
        \begin{tabular}{ |p{2.5cm} | p{4cm} |}
        \hline
             Type & Example \\
            \hline
             Conversational  &  \textcolor{red}{Well}, \textcolor{blue}{you know}, \textcolor{orange}{this is a good plan}. \\
            \hline 
             Stuttering & \textcolor{red}{Um} it was quite \textcolor{red}{fu} funny \\
            \hline
        \end{tabular}
        \caption{Examples and surface structure of disfluent utterances in conversational speech and stuttering. \textcolor{red}{Red} - Reparandum, \textcolor{blue}{Blue} - Interregnum, \textcolor{orange}{Orange} - Repair}
        
    \end{table}


\label{line_2}Data in DC is limited because of the time and resources needed to annotate data for training (\textbf{Appendix} \ref{sec:appendix_b}). Through this work\footnote{\protect\url{https://github.com/vineet2104/AdversarialTrainingForDisfluencyCorrection}}, we provide a method to create high-quality DC systems in low resource settings. Our main contributions are: 
\begin{compactenum}
\item Improving the state-of-the-art in DC in Indian languages like Bengali, Hindi and Marathi by 9.19, 5.85 and 3.40 points in F1 scores, respectively, using a deep learning framework with adversarial training on real, synthetic and unlabeled data.
\item Creating an open-source stuttering English DC corpus comprising 250 parallel sentences
\item Demonstrating that our adversarial DC model can be used for textual stuttering correction with high accuracy (87.68 F1 score)
\end{compactenum}


\section{Related work}

Approaches in DC can be categorized into noisy channel-based, parsing-based, and sequence tagging-based approaches. 
Noisy channel-based approaches rely on the following principle: a disfluent sentence Y can be obtained from a fluent sentence X by adding some noise. These models try to predict the fluent sentence X given the disfluent sentence Y (\citealp{honal_dc}; \citealp{jamshid-lou-johnson-2017-disfluency}; \citealp{10.3115/1218955.1218960}). Parsing-based approaches jointly predict the syntactic structure of the disfluent sentence along with its disfluent elements (\citealp{honnibal-johnson-2014-joint}; \citealp{jamshid-lou-johnson-2020-improving}; \citealp{rasooli-tetreault-2013-joint}; \citealp{wu-etal-2015-efficient}; \citealp{yoshikawa-etal-2016-joint}). Sequence tagging-based approaches work on the following hypothesis: every word in a disfluent sentence can be marked as fluent/disfluent. These methods work best for shorter utterances and perform optimally for real-life conversational DC (\citealp{hough_dc}; \citealp{ostendorf_dc}; \citealp{zayats_dc}). Moreover, sequence-tagging based methods require far less labeled data to perform well, compared to the other two methods. Our approach to DC focuses on treating it as a sequence tagging problem rather than a machine translation task. The objective is to accurately classify each word as either disfluent or fluent, and create fluent sentences by retaining only the fluent words. The lack of labeled data for DC in low-resource languages has prompted the use of semi-supervised methods and self-supervised techniques (\citealp{wang-etal-2018-semi}; \citealp{10.1145/3487290}). DC has also been studied as a component in speech translation systems, and thus its effect has been analyzed in improving the accuracies of machine translation models (\citealp{rao-etal-2007-improving}; \citealp{5494999}). Synthetic data generation for DC has also received attention recently. These methods infuse disfluent elements in fluent sentences to create parallel data for training (\citealp{passali-etal-2022-lard}; \citealp{saini-etal-2020-generating}). Our work is an extension of \citeauthor{kundu-etal-2022-zero} (2022), which creates the first dataset for DC in Bengali, Hindi and Marathi. We use this dataset to train our adversarial model to improve over the state-of-the-art in these languages. To the best of our knowledge, we are the first to model DC to correct stuttering ASR transcripts.

\section{Types of Disfluencies}
\label{sec:Appendix A}

There are six broad types of disfluencies encountered in real life - Filled Pause, Interjection, Discourse Marker, Repetition or Correction, False Start and Edit. Although these are common in conversational speech, stuttering speech consists mainly of Filled Pauses and Repetitions. This section describes each type of disfluency and gives some examples in English.

\vspace{5pt}

\begin{compactenum}
    \item \textbf{Filled Pauses} consist of utterances that have no semantic meaning.
    
    Example - What about the \textbf{uh} event?

    \item \textbf{Interjections} are similar to filled pauses, but their inclusion in sentences indicates affirmation or negation. 

    Example - \textbf{Ugh}, what a day it has been!

    \item \textbf{Discourse Markers} help the speaker begin a conversation or keep turn while speaking. These words do not add semantic meaning to the sentence.  

    Example - \textbf{Well}, we are going to the event.

    \item \textbf{Repetition or Correction} covers the repetition of certain words in the sentence and correcting words that were incorrectly uttered. 

    Example - If I \textbf{can't} don't go to the event today, it is not going to look good.  

    \item \textbf{False Start} occurs when previous chain of thought is abandoned, and new idea is begun. 

    Example - \textbf{Mondays dont work for me}, how about Tuesday?

    \item \textbf{Edit} refers to the set of words that are uttered to correct previous statements. 

    Example - We need \textbf{three tickets, I'm sorry}, four tickets for the flight to California. 
\end{compactenum}

\section{Architecture}

The lack of labeled data for DC is a significant hurdle to developing state-of-the-art DC systems for low-resource languages. \citeauthor{passali-etal-2022-lard} (2022), \citeauthor{saini-etal-2020-generating} (2020) and \citeauthor{kundu-etal-2022-zero} (2022) introduced data augmentation by synthesizing disfluencies in fluent sentences to generate parallel data. 
In this work, we propose a deep learning architecture that uses adversarial training to improve a BERT-based model's token classification accuracy of whether a token is disfluent or not. Our proposed architecture uses real, synthetic and unlabeled data to improve classification performance.

Our model, Seq-GAN-BERT, is inspired by \citeauthor{croce-etal-2020-gan} (2020), who first used a similar model for sentence classification. It consists of three modules: a BERT-based encoder (\citealp{devlin-etal-2019-bert}), discriminator and generator. The encoder converts the input sequence $X = (X_1,X_2,...X_n)$ into encoded vector representations ($H_{\text{real}}$). Simultaneously, the generator creates fake representations ($H_{\text{fake}}$) from Gaussian random noise (\textit{\textbf{Z}}), mimicking the real data that passes through the encoder. The discriminator aims to solve a two-pronged objective: i) predicting every word in the sentence to be  disfluent or fluent and ii) determining whether the input from the generator comes from real or fake data. 

\begin{figure}[h]
\centering
    \includegraphics[width=1.0\linewidth]{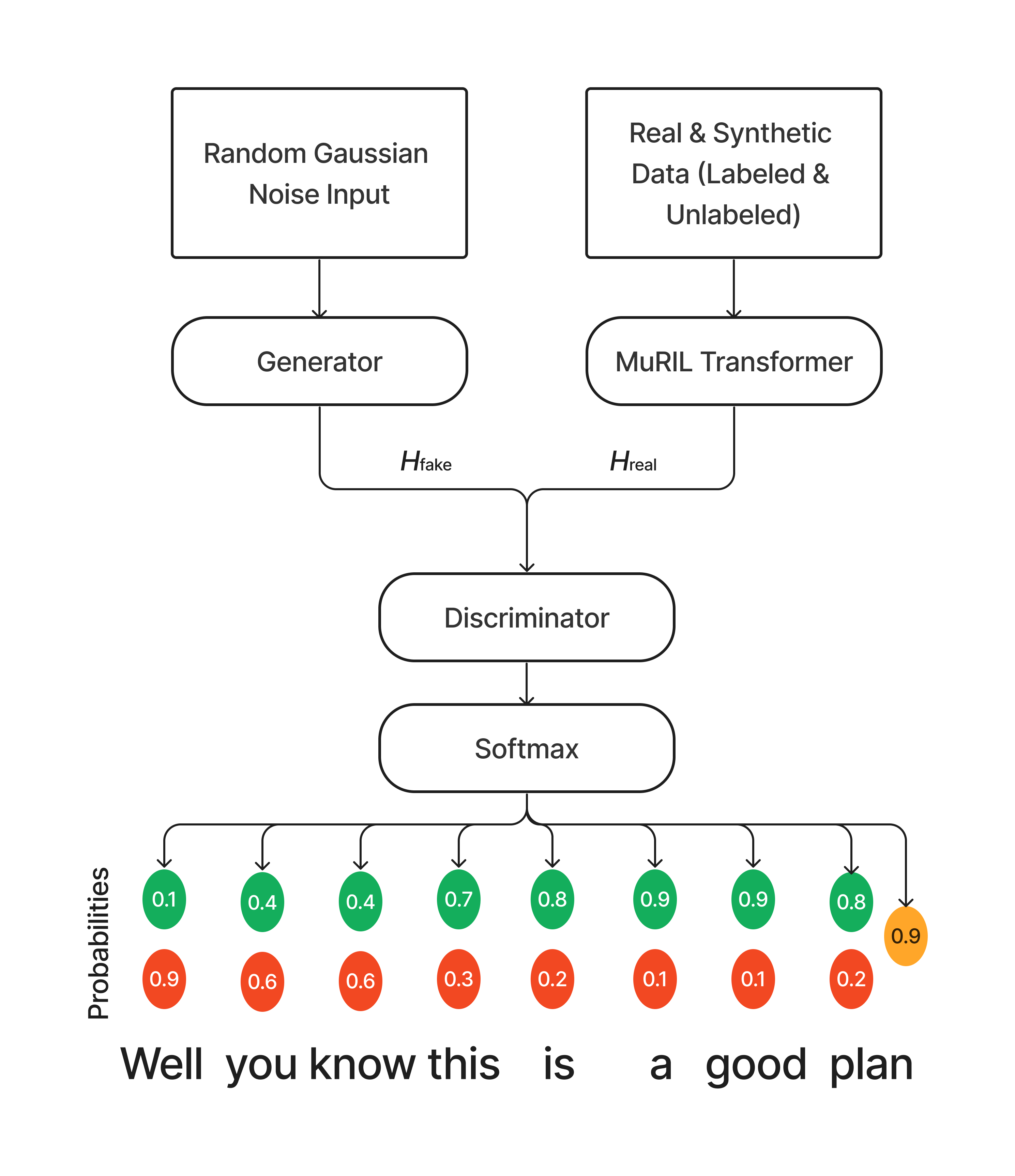}
    \caption{Architecture of the Seq-GAN-BERT model. Green nodes denote fluent class probabilities, red nodes denote disfluent class probabilities and  the orange node shows the probability of classifying a sample as real (1) or fake (0).}\label{fig:a}
\end{figure}

\subsection{Adversarial Training}
The discriminator loss comprises two loss terms. The first loss is supervised by the token classification task, while the second loss is defined by the real/fake data identification task. Such adversarial training also allows the model to use unlabeled data during training. For unlabeled samples, only the real/fake data identification task is executed. 
The generator continuously improves during training and produces fake representations that resemble actual data. The competing tasks of the generator (to create better representations to fool the discriminator) and the discriminator (to perform token classification for labeled sentences and real/fake identification) compels the MuRIL encoder to generate better representations of input sentences. The resulting high-quality representations allow the discriminator to identify disfluent words with a high accuracy. 
\vspace{-3pt}
\section{Task 1: Few Shot DC in Indian Languages}

To test our proposed architecture, we train the model on the few-shot DC task for Indian languages. The current state-of-the-art performance in Bengali, Hindi and Marathi DC is obtained by training a large multilingual transformer model using synthetic data created by injecting disfluencies in fluent sentences using rules (\citealp{kundu-etal-2022-zero}). We train our Seq-GAN-BERT model using the authors' multilingual real and synthetic data. 

\vspace{-3pt}
\subsection{Dataset}

Our dataset consists of parallel disfluent-fluent sentences in three Indian languages. We use 300, 150 and 250 real disfluent sentences in Bengali, Hindi and Marathi, respectively and generate 1000 synthetic disfluent sentences in Bengali and 500 synthetic disfluent sentences each in Hindi and Marathi each by infusing disfluent elements in fluent transcriptions using a rule-based approach (\citealp{kundu-etal-2022-zero}). The synthetic data was created such that the percentage of disfluent words across 3 languages remains constant. 

\vspace{-3pt}
\subsection{Text Processing and Training Details}
\label{subsec: process_training}

Text pre-processing is performed by removing punctuations, lower-casing and creating word-level tokens for parallel sentences. The Seq-GAN-BERT model uses a combination of labeled and unlabeled data comprising real and synthetically generated disfluent sentences in different languages. We try different combinations of monolingual and multilingual data. 
Our experiments show that the best model for Bengali uses real and synthetic Bengali sentences as labeled data and disfluent Hindi sentences as unlabeled data. The best model for Hindi uses real and synthetic Hindi sentences as labeled data and disfluent Bengali sentences as unlabeled data. The best model for Marathi uses real and synthetic Marathi sentences as labeled data and disfluent Bengali sentences as unlabeled data. The BERT-based transformer that we use as an encoder is the MuRIL model pretrained on English and many Indian languages (\citealp{khanuja_muril}). MuRIL representations for Indian languages are of superior quality compared to other multilingual Transformer-based models like mBERT (\citealp{devlin-etal-2019-bert}).   

\vspace{-3pt}

\subsection{Evaluation}
\label{subsec:4}
To evaluate our model, we train baselines for DC in zero-shot and few-shot settings. \textit{ZeroShot} is based on \citeauthor{kundu-etal-2022-zero} (2022). \textit{FewShot} is based on training MuRIL on all real and synthetic data available in the chosen language, along with labeled data in a related Indian language (for Bengali, either Hindi or Marathi can act as a related Indian language). \textit{FewShotAdv} is the Seq-GAN-BERT model without any unlabeled data. Although models like BiLSTM-CRF have been as alternatives to transformers for sequence tagging, direct finetuning often performs better (\citealp{Ghosh2022SpanCW}). Performance of DC systems is usually measured with F1 scores (\citealp{ferguson-etal-2015-disfluency}; \citealp{honnibal-johnson-2014-joint}; \citealp{jamshid-lou-johnson-2017-disfluency}). 
Table 3 shows the comparison of various baselines against our model. 




\begin{table}[h]
        \centering
        \begin{tabular}{ |c | c | c | c | c |}
        \hline
            Lang & Model & P & R & F1 \\
            \hline
            Bn & ZeroShot & 93.06 & 62.18 & 74.55 \\
            & FewShot & 66.37 & 68.20 & 67.27 \\
            & FewShotAdv & 84.00 & 78.93 & 81.39 \\
            & Our model & 87.57 & 80.23 & \textbf{83.74} \\
            \hline 
            Hi & ZeroShot &  85.38 & 79.41 & 82.29 \\
            & FewShot &  82.99 & 81.33 & 82.15 \\
            & FewShotAdv &  88.15 & 83.14 & 85.57 \\
            & Our model & 89.83 & 86.51 & \textbf{88.14} \\
            \hline
            Mr & ZeroShot & 87.39 & 61.26 & 72.03 \\
            & FewShot & 82.00 & 60.00 & 69.30 \\
            & FewShotAdv & 84.21 & 64.21 & 72.86 \\
            & Our model & 85.34 & 67.58 & \textbf{75.43} \\
        \hline
        \end{tabular}
        \caption{Comparing the performance of baselines and our model on DC across Bengali (Bn), Hindi (Hi) and Marathi (Mr); ZeroShot - Monolingual supervised training, FewShot - Multilingual supervised training, FewShotAdv - Adversarial training without unlabeled data, Our model - Multilingual adversarial training with unlabeled data; P = Precision, R = Recall}
        
    \end{table}
    
 Our model, Seq-GAN-BERT with unlabeled sentences, performs better than the other baselines and establishes a new state-of-the-art for DC in Bengali, Hindi and Marathi. Our model benefits from adversarial training using both unlabeled data and multilingual training. 
 The observed precision and recall scores of these models during testing show that without adversarial training, the model performs with high precision but low recall. However, with adversarial training, the model improves its recall without compromising much on precision. The zero-shot model (without adversarial training) classifies less words as disfluent but at a high accuracy, whereas the few-shot model (with adversarial training) correctly classifies more words as disfluent. 

\section{Task 2: Stuttering DC in English}

We have already shown how our proposed architecture learns better semantic representations for DC using small amounts of manually annotated labeled data.  In this section, we present a similar experiment in Stuttering DC (SDC). We define SDC as the task of removing disfluent elements in spoken utterances that are caused by stuttering speech impairment. Since this is the first attempt to model stuttering correction as disfluency removal, we make our version of the existing dataset for stuttering publicly available for research purposes and provide various baseline comparisons. We show that our model generalizes well for this task and is able to remove disfluent elements in stuttering speech.  

\vspace{-5pt}

\subsection{Dataset}

The UCLASS dataset is created by transcribing audio interviews of 14 anonymous teenagers who stutter and consists of two released versions (\citealp{howell_effectiveness_2004}). Both versions of this corpus are available for free download and research. We create 250 disfluent-fluent parallel sentences from the available transcripts of such utterances. The dataset is released here\footnote{\protect\url{https://github.com/vineet2104/AdversarialTrainingForDisfluencyCorrection}}. 

\subsection{Processing \& Training}

We follow the same steps as before (section  \ref{subsec: process_training}). Stuttered syllables are represented in the text, separated by a space delimiter and treated as a disfluent term. This gold-standard dataset is split into 150 sentences for training and 100 sentences for testing. The training sentences are used as labeled data for the model and unlabeled data from Switchboard (\citealp{Godfrey1992SWITCHBOARDTS}) or \citeauthor{kundu-etal-2022-zero} (2022) is used to facilitate multilingual training. Our model performs best when we use synthetic Bengali disfluent sentences as unlabeled data. 

\subsection{Evaluation}

We use five baselines to evaluate our model's performance. \textit{SupervisedGold} uses the gold standard data and trains the MuRIL model for token classification. \textit{SupervisedGoldSWBD and SupervisedGoldLARD} uses a combination of the gold standard dataset along with 1000 disfluent sentences from the Switchboard corpus and LARD dataset (\citealp{passali-etal-2021-towards}). \textit{AdversarialSWBD and AdversarialLARD} uses the Seq-GAN-BERT to train on a combination of labeled sentences from gold standard corpus and unlabeled sentences from the Switchboard corpus and LARD dataset. Table 4 displays our results averaged over multiple seeds.

Our model outperforms all baselines. Improvement over \textit{AdversarialLARD} shows the benefit of multilingual training. We also used synthetic Hindi or Marathi data while training, but achieved lower scores than the \textit{AdversarialLARD} baseline.  

\begin{table}
        \centering
        
        \begin{tabular}{ | c | c | c | c |}
        \hline
            Model & P & R & F1 \\
            \hline
            SupervisedGold & 89.11 & 78.08 & 83.23 \\
            \hline
            SupervisedGoldSWBD & 87.34 & 86.50 & 86.92 \\
            \hline
            SupervisedGoldLARD & 74.58 & 86.33 & 80.02 \\
            \hline
            AdversarialSWBD & 85.76 & 84.17 & 84.96 \\
            \hline
            AdversarialLARD & 86.21 & 84.82 & 85.51 \\
            \hline
            Our model & 87.26 & 88.10 & \textbf{87.68} \\

        \hline
        \end{tabular}
        \caption{Comparing baselines and our model for English stuttering DC; SupervisedGold - Supervised training on gold standard dataset, SupervisedGoldSWBD and SupervisedGoldLARD - Supervised training on gold standard dataset and DC data, AdversarialSWBD and AdversarialLARD - Adversarial training without unlabeled data, Our model - Multilingual Adversarial training with unlabeled data; P = Precision, R = Recall}
        
    \end{table}

\vspace{-3.5pt}
\noindent \textbf{Summary of results:} In this paper, we evaluate our proposed architecture for low-resource DC using two tasks: 1) DC in Indian languages and 2) Stuttering DC in English. Our model outperforms competitive baselines across both these tasks establishing a new state-of-the-art for Indian languages DC. The adversarial training in our model improves the representations of a BERT-based encoder for disfluent/fluent classification. We show that multilingual training benefits such tasks as the generator is trained to create better representations of fake data to fool the discriminator. 


\vspace{-3.5pt}

\section{Conclusion}
\vspace{-2pt}
Adversarial training using unlabeled data can benefit disfluency correction when we have limited amounts of labeled data. Our proposed model can also be used to correct stuttering in ASR transcripts with high accuracy.



Future work lies in integrating speech recognition models like Whisper\footnote{\protect\url{https://cdn.openai.com/papers/whisper.pdf}} or wav2vec 2.0 (\citealp{baevski_whisper}) to create end-to-end speech-driven DC models. It will also be insightful to see how this model transfers to other low-resource languages with different linguistic properties. 

\section{Acknowledgements}

We would like to thank the anonymous reviewers and area chairs for their suggestions to strengthen the paper. This work was done as part of the Bahubhashak Pilot Project on Speech to Speech Machine
Translation under the umbrella of National Language Technology Mission of Ministry of Electronics and IT, Govt. of India.  We would also like to thank Nikhil Saini for valuable discussions during the course of this project.

\section{Limitations}

There are two main limitations of our work. Firstly, since there are no known baselines for Indian language DC except \citeauthor{kundu-etal-2022-zero} (2022), other architectures might perform better than our model. Our claim that Seq-GAN-BERT tries to maximize the information gained from unlabeled sentences is supported by superior performance over baselines defined in this work and other related models. Secondly, due to the lack of good quality labeled datasets, our test sets contained only 100 sentences. However, we believe that the consistency of our high-performing models across languages and multiple seeded experiments presents a positive sign for DC in low-resource settings. 

\section{Ethics Statement}
The aim of our work was to design an adversarial training-enabled token classification system that is able to correctly remove disfluencies in text. The datasets used in this work are publicly available and we have cited the sources of all the datasets that we have used. 

\newpage





\bibliography{anthology,custom}
\bibliographystyle{acl_natbib}

\end{document}